\documentclass[10pt,twocolumn,letterpaper]{article}

\usepackage{cvpr}
\usepackage{times}
\usepackage{epsfig}
\usepackage{graphicx}
\usepackage{amsmath}
\usepackage{amssymb}

\usepackage{array}
\usepackage{multirow}

\usepackage[pagebackref=true,breaklinks=true,letterpaper=true,colorlinks,bookmarks=false]{hyperref}

\cvprfinalcopy 

\def\cvprPaperID{8980} 

\ifcvprfinal\fi
\begin{document}

\title{PropagationNet: Propagate Points to Curve to Learn Structure Information}

\author{Xiehe~Huang\textsuperscript{\rm 1, 2}\and Weihong~Deng\textsuperscript{\rm 1}\\
\and Haifeng~Shen\textsuperscript{\rm 2} \and Xiubao~Zhang\textsuperscript{\rm 2} \and Jieping~Ye\textsuperscript{\rm 2} \\
\textsuperscript{\rm 1}Pattern Recognition \& Intelligent System Laboratory, School of Information and \\
Communication Engineering, Beijing University of Posts and Telecommunications\\
{\tt\small \{xiehe.huang, whdeng\}@bupt.edu.cn}\\
\textsuperscript{\rm 2}AI Labs, DiDi Chuxing\\
{\tt\small \{huangxiehe\_i, shenhaifeng, zhangxiubao, yejieping\}@didiglobal.com}
}
\maketitle
\thispagestyle{empty}

\begin{abstract}
	Deep learning technique has dramatically boosted the performance of face alignment algorithms. However, due to large variability and lack of samples, the alignment problem in unconstrained situations, \eg large head poses, exaggerated expression, and uneven illumination, is still largely unsolved. In this paper, we explore the instincts and reasons behind our two proposals, \ie Propagation Module and Focal Wing Loss, to tackle the problem. Concretely, we present a novel structure-infused face alignment algorithm based on heatmap regression via propagating landmark heatmaps to boundary heatmaps, which provide structure information for further attention map generation. Moreover, we propose a Focal Wing Loss for mining and emphasizing the difficult samples under in-the-wild condition. In addition, we adopt methods like CoordConv and Anti-aliased CNN from other fields that address the shift-variance problem of CNN for face alignment. When implementing extensive experiments on different benchmarks, \ie WFLW, 300W, and COFW, our method outperforms state-of-the-arts by a significant margin. Our proposed approach achieves 4.05\% mean error on WFLW, 2.93\% mean error on 300W full-set, and 3.71\% mean error on COFW.
\end{abstract}

\section{Introduction}

	Face alignment, aimed at localizing facial landmark, plays an essential role in many face analysis applications, \eg face verification and recognition \cite{Wang2018DeepFR}, face morphing \cite{hassner2015effective}, expression recognition \cite{Li2018DeepFE}, and 3D face reconstruction \cite{dou2017end}. Recent years have witnessed the constant emergence of fancy face alignment algorithms with considerable performance on various datasets. Yet face alignment in unconstrained situations, \eg large head pose, exaggerated expression, and uneven illumination, has plagued researchers over the years. Among many other factors, we attribute the mentioned challenges to the disability of CNN to learn facial structure information: if a CNN is enabled to extract the structure of a face in an image, then it can predict facial landmarks more accurately since even the occluded parts of a face, for instance, can be inferenced through the shape of the face. This is also the intention of ASM \cite{cootes1995active}'s designers.

\begin{figure}[t]
\centering
\includegraphics[width=0.48\textwidth]{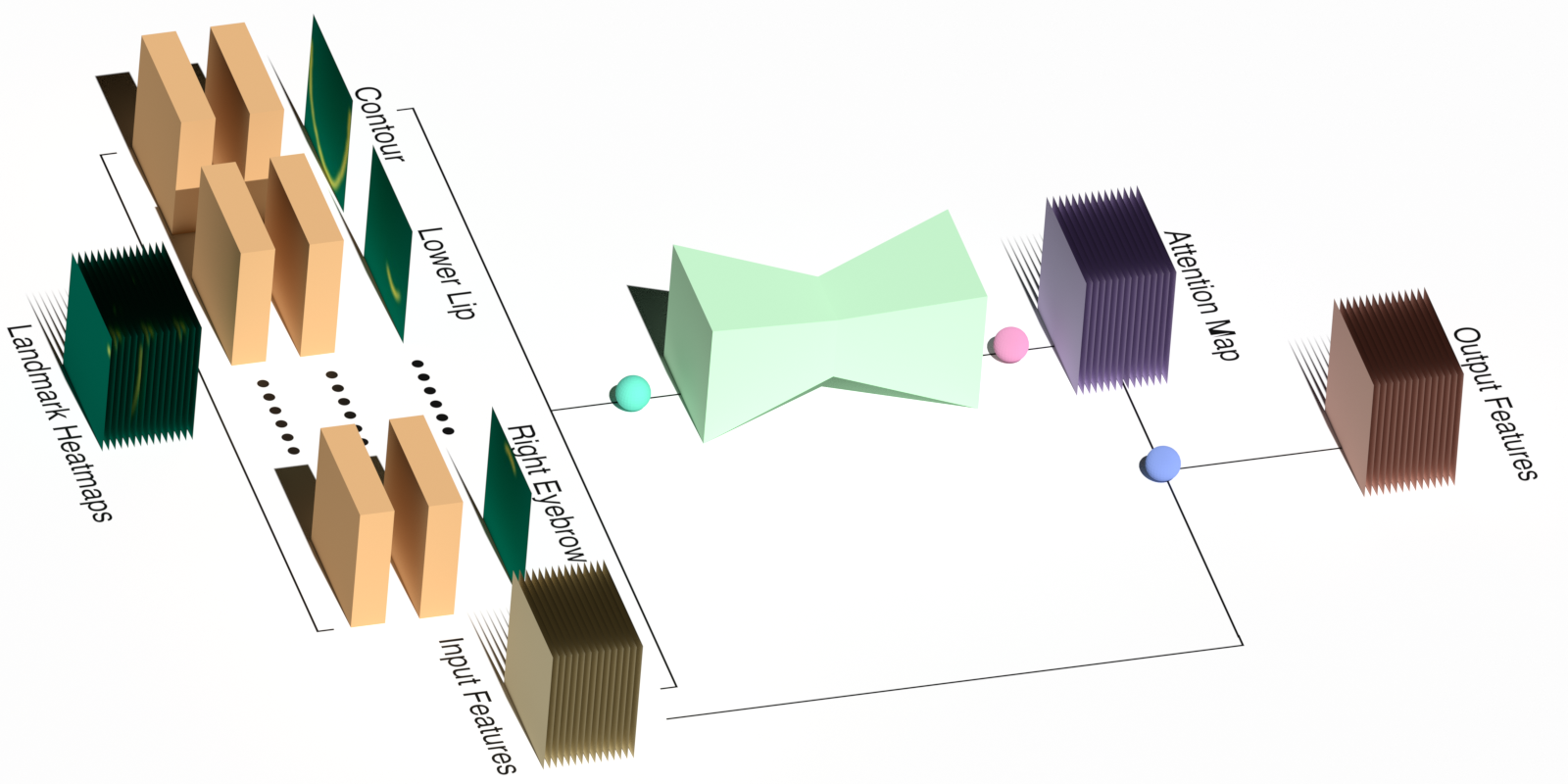} 
\caption{Building blocks of our propagation module. Landmark heatmaps are input to multiple convolutional operations, then concatenated with the output feature maps of last hourglass module, together processed by a two-stage hourglass module, and finally normalized by a sigmoid layer to form an attention map that is imposed on the feature maps.}
\label{Block}
\end{figure}

\begin{figure}[h]
\centering
\includegraphics[width=0.4\textwidth]{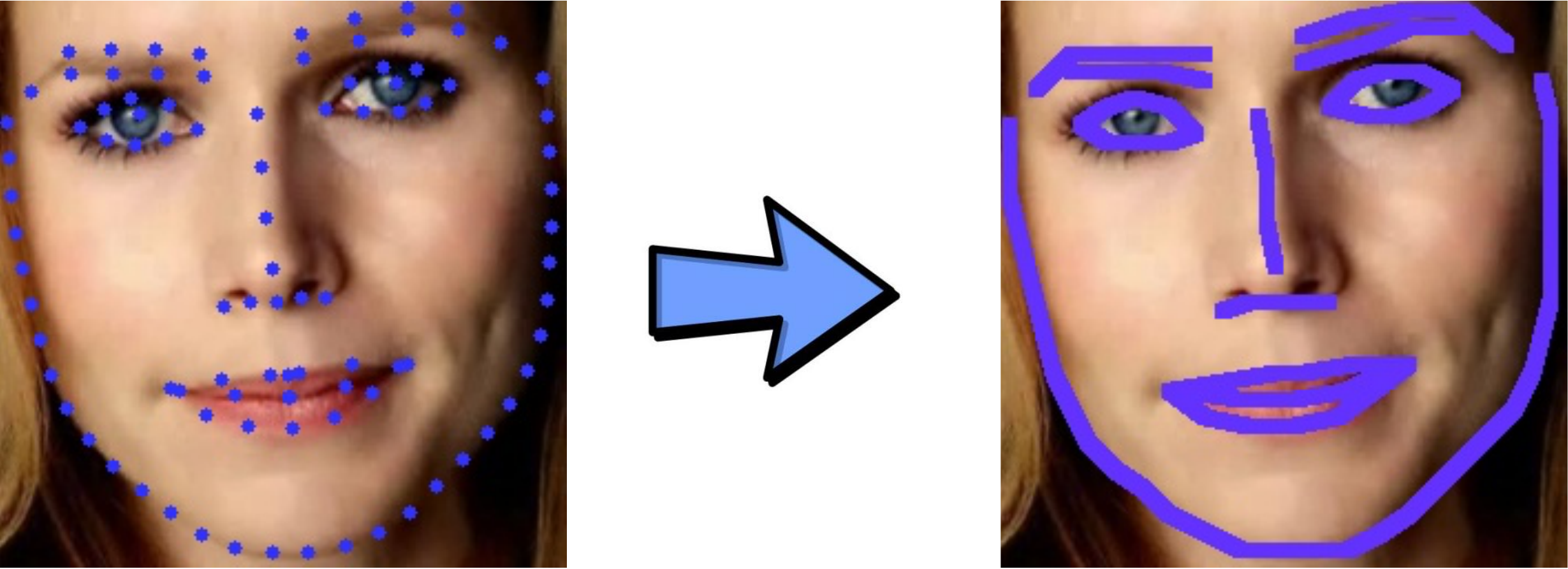} 
\caption{Landmarks are connected to make several boundaries.}
\label{Boundary}
\end{figure}

	What exactly is structural information? In our work, we deem it to be the statistical mean of landmark coordinates. Perhaps with high variance (such as different head poses), landmark coordinates are still subject to some distribution due to the relative non-transformability of a face shape. In order for a CNN to learn the information, we represent it as facial boundary in this paper (see Fig. \ref{Boundary}) following Wu \etal \cite{wu2018look}. A facial boundary could be the jawline, or the outer contour of a face. Or it could be the edge surrounding a mouth. These boundaries are commonly annotated with a series of points by available datasets due to their difficulty with modeling a line.

	In this paper, we propose and implement 3 creative ideas to learn the structural information, , \ie Propagation Module, Focal Wing Loss, and Multi-view Hourglass.

	Wu \etal produce facial boundaries out of a separate GAN (Generative Adversarial Network) generator. Specifically, Wu \etal connected the landmarks to make a blurred line and specified it as ground truth for future training. Unlike their fashion of using an independent CNN to generate facial boundaries, we devise a \textbf{Propagation Module} to do this job and incorporate it within our network architecture, in an attempt to substitute a deeper and larger CNN with a computation-efficient Propagation Module. In addition to the module's embeddability in a CNN, what is more important is that boundary heatmaps are naturally connected with landmark heatmaps. For this reason, it is intuitive for us to use a series of convolution operations to model the connection and propagate a certain number of landmarks (points) to a boundary (a curve). Hence we term the module as Propagation Module.

\begin{figure}[h]
\centering
\includegraphics[width=0.48\textwidth]{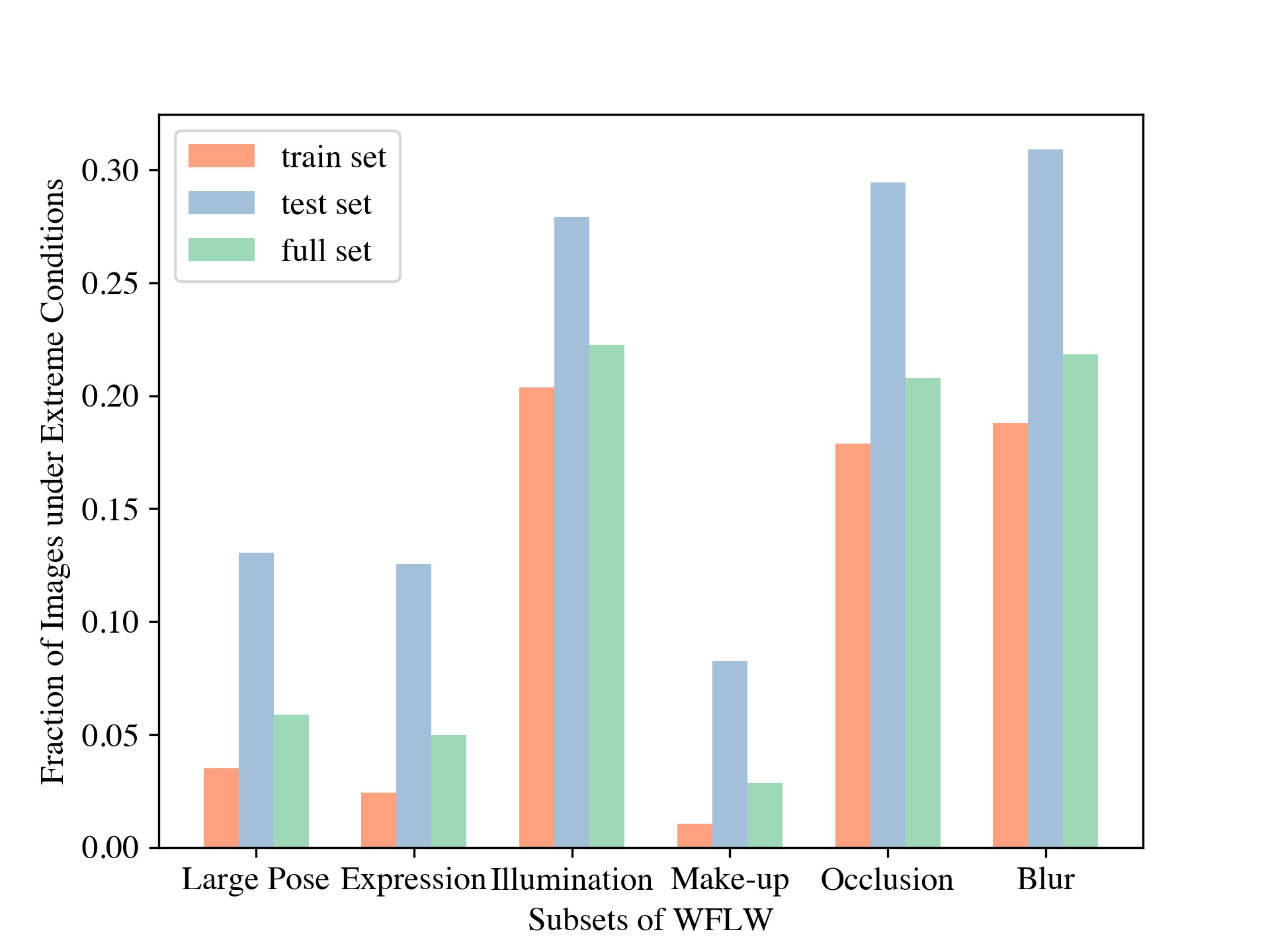} 
\caption{Fractions of images under extreme conditions for different sets. Extreme conditions include large head pose, exaggerated expression, non-uniform illumination, unrecognizable make-up, object occlusion, and blurry shot.}
\label{stat}
\end{figure}

	Data imbalance is a common issue in many fields of AI and so is the case in face alignment community. The structures of a face varies under in-the-wild conditions. For example, the jawline is less widely open when a face is in profile position than when the face is shown frontal. However, the ratio of data under these two conditions is not actually anywhere near $1:1$, where the number of frontal images is the same as that of profile ones. As illustrated in Fig. \ref{stat}, the fractions of images under extreme conditions are rather low, all under 30\% across both train set and test set. On the other hand, the fractions on train set deviates largely from those on test set, which means a learned feature adapted to train set can misguide the neural network to make a wrong prediction. This potential non-universal feature, therefore, necessitates a better design of loss function. Based on the primitive AWing \cite{wang2019adaptive}, we propose a \textbf{Focal Wing Loss}, which dynamically adjusts the penalty for incorrect prediction and gears the loss weight (thus the learning rate) for each sample in every batch during training. This indicates that our training process pay attention evenly to both hard-to-learn structures and easy-to-learn ones, so we refer to the loss function as Focal Wing Loss.

	Modern-day convolutional neural networks are usually believed to be shift invariant, and so is the stacked hourglass used in our work. Nevertheless, researchers have come to realize the underlying shift variance brought by the introduction of pooling layer, \eg max pooling and average pooling. To resolve this translation-variance, Zhang \cite{zhang2019making} provided the solution of Anti-aliased CNN, which emulates the traditional signal processing method of anti-aliasing and apply it before every downsampling operation, such as pooling and strided convolution. In our task, we cannot afford to lose structural information when applying pooling layer, so we incorporate Anti-aliased CNN in a special hourglass and name it \textbf{Multi-view Hourglass}.

	In conclusion, our main contribution include:
	\begin{itemize}
		\item creating a novel propagation module to seamlessly connect landmark heatmaps with boundary heatmaps, a module that can be naturally built into stacked hourglass model.
		\item devising a loss function termed as Focal Wing Loss to dynamically assign loss weight to a particular sample and tackle data imbalance.
		\item introducing Anti-aliased CNN from other fields and integrating them within our Multi-View Hourglass Module to add shift equivariance and coordinate information to our network.
		\item implementing extensive experiments on various datasets as well as ablation studies about the mentioned methods.
	\end{itemize}

\section{Related Work}

\begin{figure}[t]
\centering
\includegraphics[width=0.48\textwidth]{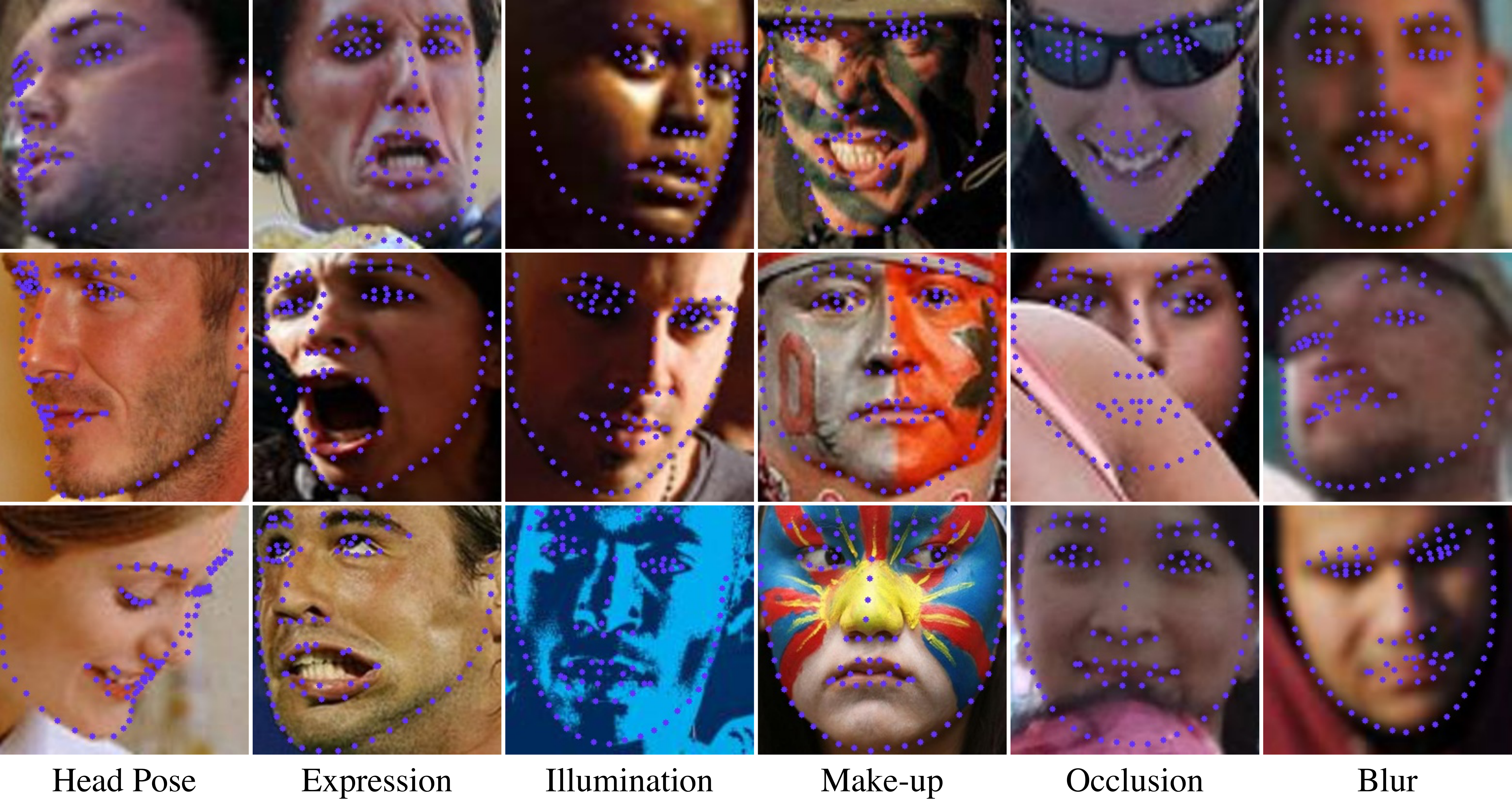} 
\caption{Sample results on WFLW testset. Each column comes from a subset of WFLW, including large head pose, expression, illumination, make-up, occlusion, and blur.}
\label{ResultWFLW}
\end{figure}

	Recently, interest of face alignment community has been largely centering on two mainstream approaches, \ie coordinate regression and heatmap regression, with various model designs. \textbf{Heatmap regression models}, based on fully convolution network (FCN), output a heatmap for each landmark and try to maintain structure information throughout the whole network, therefore, to some extent, dwarfing coordinate regression models in their state-of-the-art performance. MHM \cite{deng2019joint}, one of those heatmap regression models, implements face detection and face alignment consecutively and leverages stacked hourglass model to predict landmark heatmaps. AWing \cite{wang2019adaptive}, yet another heatmap regression model, modifies L1 loss to derive so-called adaptive wing loss and proves its superiority in CNN-based facial landmark localization. What is common among these 2 models is their adoption of stacked hourglass network. Stacked hourglass model stands out among all FCNs in the field of landmark detection since its debut in \cite{newell2016stacked} for human pose estimation. Its prevalence can be attributed to its repeated bottom-up, top-down processing that allows for capturing information across all scales of an input image.

	First raised by Wu \etal \cite{wu2018look} and later popularized by such researchers as Wang \etal \cite{wang2019adaptive}, \textbf{facial boundary} identifies geometry structure of human face and therefore can infuse a network with prior knowledge, be it used for attention mechanism (as in the case of LAB \cite{wu2018look}) or for generation of boundary coordinate map (as in the case of AWing \cite{wang2019adaptive}). In the former scenario, LAB first utilizes a stacked hourglass model to generate facial boundary map and then incorporates the boundary map to a regression network via feature map fusion. In the latter scenario, AWing encodes boundary prediction as a mask on x-y coordinates and consequently produces two additional feature maps for follow-on convolution. Different from both of them, we generate the boundary heatmap with only several convolution operations instead of a complicated CNN.

\begin{figure}[t]
\centering
\includegraphics[width=0.48\textwidth]{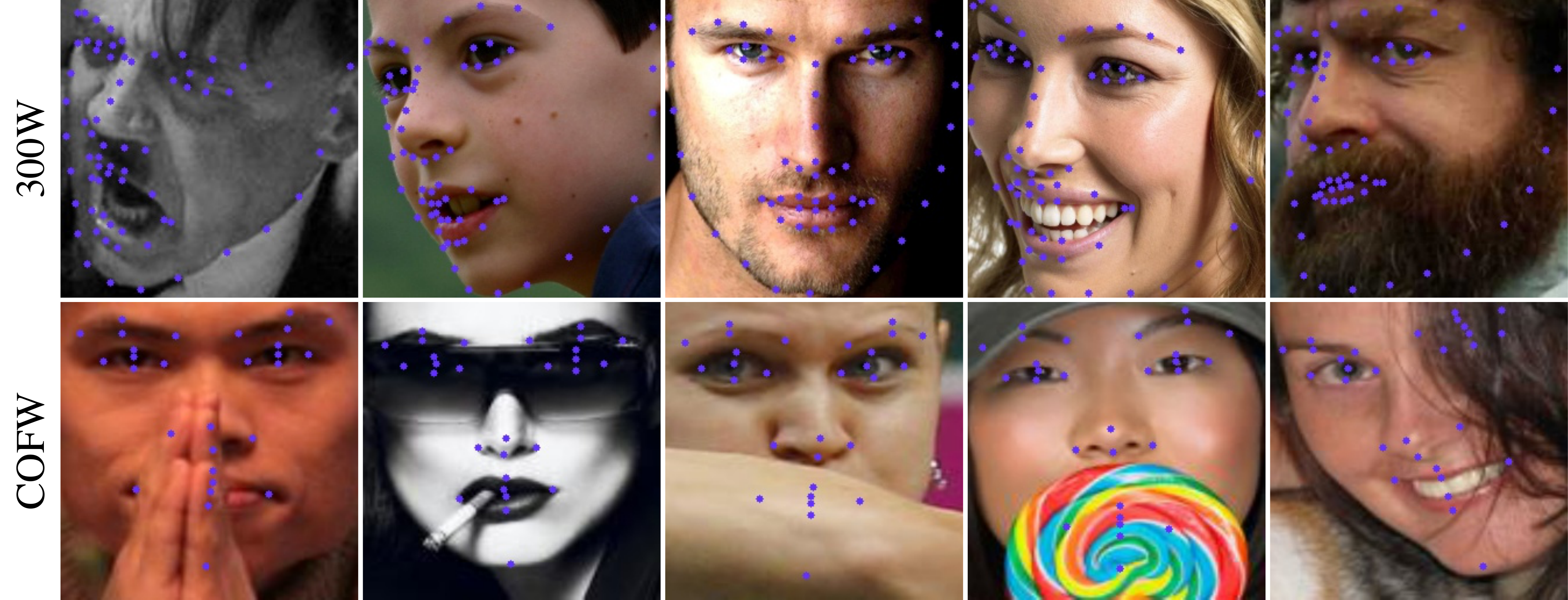} 
\caption{Sample results on 300W and COFW testsets. Each row demonstrates samples from each dataset.}
\label{ResultW}
\end{figure}

	\textbf{Attention mechanism} has enjoyed great popularity in computer vision because the extra ``attention'' brought by it can guide a CNN to learn worthable features and focus on them. In our work, we want our model to focus more on the area of boundary so it can inference landmark more accurately based on the position of boundary. Unlike LAB \cite{wu2018look}' way of using a ResNet-block-based hourglass to generate attention map, we adopt a multi-view hourglass which can maintain structual information throughout the whole process. Specifically, we incorporate Hierarchical, Parallel \& Multi-Scale block \cite{bulat2017binarized} to add more sizes of receptive fields and Anti-aliased CNN \cite{zhang2019making} to improve shift invariance. Larger size of receptive field means that our model can ``behold'' the whole structure of a face, whereas shift invariance means that our model can still predict boundary heatmaps correctly even if the correspondent face image is shifted a little bit. Moreover, we do not have to downsample boundary heatmaps every time they are fed into the next hourglass whereas LAB does. This is because we do not want to lose boundary information via downsampling.

	CNN-based localization models have long been trained with take-away \textbf{loss functions}, \eg L1, L2, and smooth L1. These loss functions are indeed useful in common scenarios. Feng \etal \cite{feng2018wing}, however, contends that L2 is sensitive to outliers and therefore dwarfed by L1. In order to make their model pay more attention to small and medium range errors, they modify the L1 loss to create Wing Loss which is more effective in landmark coordinates regression models. Based on Wing Loss, Wang \etal \cite{wang2019adaptive} further bring in adaptiveness to the loss function because they believe ``influence" (a concept lent from robust statistics) should be proportional to gradient and balance all errors. The Adptive Wing Loss they created is proven to be more effective in heatmap regression models.

\section{Approach}

\begin{figure*}[t]
\centering
\includegraphics[width=1\textwidth]{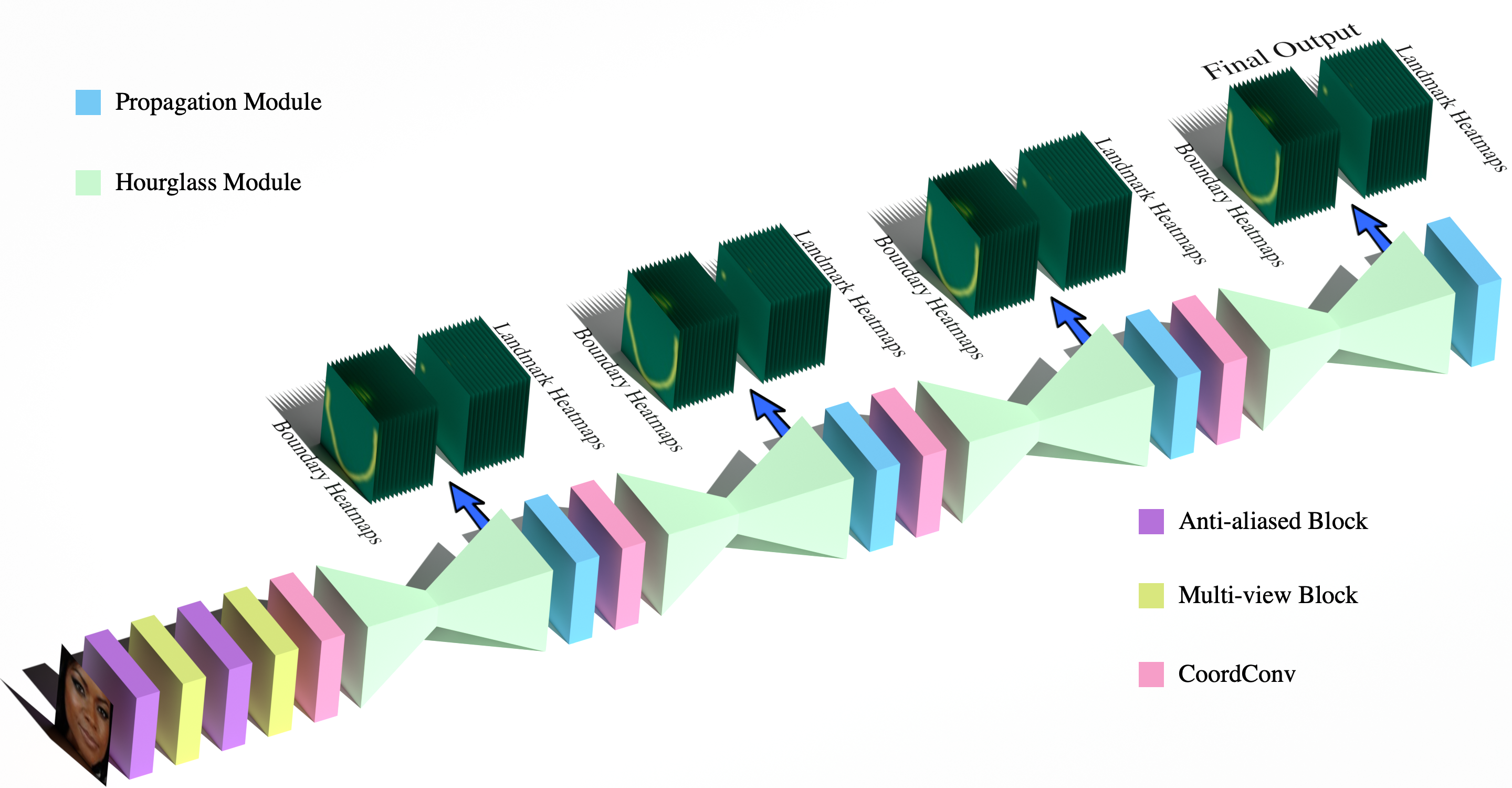} 
\caption{Overview of our PropogationNet architecture. RGB images are first processed through a series of basic feature extractors, and then fed into several hourglass modules followed by a relation block that outputs the boundary heatmaps.}
\label{architecture}
\end{figure*}

	Based on stacked HG design from Bulat \etal \cite{bulat2017binarized}, our model further integrates it with Propagation Module, anti-aliased block, and CoordConv. Each hourglass outputs feature maps for the following hourglass and landmark heatmaps supervised with ground truth labels. Next in line is a Propagation Module, which generates boundary heatmaps and outputs the feature maps for the follow-on hourglass. This overall process is visualized in Fig. \ref{architecture}.

\begin{table}[h]
\caption{Evaluation of PropagationNet and other state-of-the-arts on COFW testset.}\label{COFW}
\centering
\begin{tabular}{|c|c|c|}
\hline
Method & NME & $\text{FR}_\text{10\%}$ \\
\hline
Human \cite{burgos2013robust} & 5.60 & - \\
RCPR \cite{burgos2013robust} & 8.50 & 20.00 \\
TCDCN \cite{zhang2014facial} & 8.05 & - \\
DAC-CSR \cite{feng2017dynamic} & 6.03 & 4.73 \\
Wing \cite{feng2018wing} & 5.44 & 3.75 \\
Awing \cite{wang2019adaptive} & 4.94 & 0.99 \\
LAB \cite{wu2018look} & 3.92 & 0.39 \\
\hline
\textbf{PropNet(Ours)} & \textbf{3.71} & \textbf{0.20} \\
\hline
\end{tabular}
\end{table}

\subsection{Landmark-Boundary Propagation Module}

	Inspired by attention mechanism, the Landmark-Boundary Propagation Module aims to force the network to pay more ``attention" to the boundary area in order to make more accurate prediction of landmark heatmap. To achieve this goal, it first employs an array of convolution operations to transform landmark heatmaps to boundary heatmaps. These operations basically attempt to learn how to translate landmark heatmaps and combine the trajectories to make boundary heatmaps. Every boundary heatmap is generated via a set of several $7\times7$ convolution operations. Then it concatenates boundary heatmaps and the feature maps from its anterior hourglass module and feeds them into a two-stage hourglass module in order to yield the attention map. Finally, it enhances the feature maps with attention map and transports those feature maps to its posterior hourglass. This process is visualized in Fig. \ref{Block}.

	During training, the generation of boundary heatmaps is supervised by ground truth heatmaps. As for how to produce ground truth heatmaps, we simply link landmarks together with straight lines and apply Gaussian blurring filter. Each boundary has its semantic meanings. As depicted in Fig. \ref{Boundary}, landmarks lying on the jawline are connected to make contour boundary, those denoting the lower lip are connected to make another boundary, and so forth. In total, we obtain $M=15$ boundary heatmaps.

\subsection{Focal Wing Loss}
	Adaptive wing loss \cite{wang2019adaptive} is derived from wing loss \cite{feng2018wing} and is basically a variant of smooth L1 loss except that the smooth quadratic curve is replaced by a logarithmic curve. It is piecewise-defined as Eq. (\ref{wing}), where $A=\omega\left(\alpha-y\right)\left(\theta/\epsilon\right)^{\alpha-y-1}/{\left(1+\left(\theta/\epsilon\right)^{\alpha-y}\right)}/\epsilon$ and $\Omega=\theta A -\omega ln\left(1+\left(\theta/\epsilon\right)^{\alpha-y}\right)$ are defined to make the loss function continuous and smooth at $\left|y-\hat{y}\right|=\theta$ and $\omega$, $\theta$, $\alpha$, and $\epsilon$ are hyper-parameters that affect the none-L1 range and the gradient between it.

\begin{equation}\label{wing}
AWing\left(x\right) =
\begin{cases}
\omega ln\left(1+|\frac{y-\hat{y}}{\epsilon}|^{\alpha-y}\right), & \left|y-\hat{y}\right|<\theta \\
A\left|y-\hat{y}\right|-\Omega , & \left|y-\hat{y}\right|\geq\theta
\end{cases}
\end{equation}

\begin{table}
\caption{Evaluation of PropagationNet and other state-of-the-arts on 300W testset.}\label{300W}
\centering
\begin{tabular}{|c|c|c|c|}
\hline
\multirow{2}{*}{Method} & Common & Challenging & \multirow{2}{*}{Fullset} \\
~ & Subset & Subset & ~ \\
\hline
\multicolumn{4}{|c|}{Inter-pupil Normalization} \\
\hline
CFAN \cite{zhang2014coarse} & 5.50 & 16.78 & 7.69 \\
SDM \cite{xiong2013supervised} & 5.57 & 15.40 & 7.50 \\
LBF \cite{ren2014face}& 4.95 & 11.98 & 6.32 \\
CFSS \cite{zhu2015face}& 4.73 & 9.98 & 5.76 \\
TCDCN \cite{zhang2015learning}&  4.80 & 8.60 & 5.54 \\
MDM \cite{trigeorgis2016mnemonic}& 4.83 & 10.14 & 5.88 \\
RAR \cite{xiao2016robust}& 4.12 & 8.35 & 4.94 \\
DVLN \cite{wu2015robust}& 3.94 & 7.62 & 4.66 \\
TSR \cite{lv2017deep}& 4.36 & 7.56 & 4.99 \\
DSRN \cite{miao2018direct}& 4.12 & 9.68 & 5.21 \\
LAB \cite{wu2018look}& 4.20 & 7.41 & 4.92 \\
$\text{RCN}^+$(L+ELT) \cite{honari2018improving} & 4.20 & 7.78 & 4.90 \\
DCFE \cite{valle2018deeply}& 3.83 & 7.54 &  4.55 \\ 
Wing \cite{feng2018wing}& \textbf{3.27} & 7.18 & \textbf{4.04} \\
AWing \cite{wang2019adaptive}& 3.77 & 6.52 & 4.31 \\
\hline
\textbf{PropNet(Ours)} & 3.70 & \textbf{5.75} & 4.10\\
\hline
\multicolumn{4}{|c|}{Inter-ocular Normalization} \\
\hline
PCD-CNN \cite{kumar2018disentangling}& 3.67 &  7.62 &  4.44 \\
CPM+SBR \cite{dong2018style}& 3.28 &  7.58 &  4.10 \\
SAN \cite{dong2018style}& 3.34  & 6.60 &  3.98 \\
LAB \cite{wu2018look}& 2.98 &  5.19 &  3.49 \\
DU-Net \cite{tang2018quantized}& 2.90 &  5.15 &  3.35\\
AWing \cite{wang2019adaptive}& 2.72 &  4.52 &  3.07 \\
\hline
\textbf{PropNet(Ours)} & \textbf{2.67} & \textbf{3.99} & \textbf{2.93}\\
\hline
\end{tabular}
\end{table}

\begin{table*}
\caption{Evaluation of PropagationNet and other state-of-the-arts on WFLW testset and its subsets.}\label{WFLW}
\centering
\begin{tabular}{|c|c|c|c|c|c|c|c|c|}
\hline
\multirow{2}{*}{Metric} & \multirow{2}{*}{Method} & \multirow{2}{*}{Testset} & Pose & Expression & Illumination &  Make-up & Occlusion  & Blur\\
~ & ~ & ~ & Subset & Subset & Subset & Subset & Subset & Subset \\
\hline
\multirow{7}{*}{NME (\%)} & ESR \cite{cao2014face} & 11.13 & 25.88 & 11.47 & 10.49 & 11.05 & 13.75 & 12.20 \\
~ & SDM \cite{xiong2013supervised} & 10.29 & 24.10 & 11.45 & 9.32 & 9.38 & 13.03 & 11.28 \\
~ & CFSS \cite{zhu2015face} & 9.07 & 21.36 & 10.09 & 8.30 & 8.74 & 11.76 & 9.96 \\
~ & DVLN \cite{wu2017leveraging} & 6.08 & 11.54 & 6.78 & 5.73 & 5.98 & 7.33 & 6.88 \\
~ & LAB \cite{wu2018look} & 5.27 & 10.24 & 5.51 & 5.23 & 5.15 & 6.79 & 6.12 \\
~ & Wing \cite{feng2018wing} & 5.11 & 8.75 & 5.36 & 4.93 & 5.41 & 6.37 & 5.81 \\
~ & \textbf{PropNet(Ours)} & \textbf{4.05} & \textbf{6.92} & \textbf{3.87} & \textbf{4.07} & \textbf{3.76} & \textbf{4.58} & \textbf{4.36} \\
\hline
\multirow{7}{*}{$\text{FR}_\text{10\%}$ (\%)} & ESR \cite{cao2014face} & 35.24 & 90.18 & 42.04 & 30.80 & 38.84 & 47.28 & 41.40 \\
~ & SDM \cite{xiong2013supervised} & 29.40 & 84.36 & 33.44 & 26.22 & 27.67 & 41.85 & 35.32 \\
~ & CFSS \cite{zhu2015face} & 20.56 & 66.26 & 23.25 & 17.34 & 21.84 & 32.88 & 23.67 \\
~ & DVLN \cite{wu2017leveraging} & 10.84 & 46.93 & 11.15 & 7.31 & 11.65 & 16.30 & 13.71 \\
~ & LAB \cite{wu2018look} & 7.56 & 28.83 & 6.37 & 6.73 & 7.77 & 13.72 & 10.74 \\
~ & Wing \cite{feng2018wing} & 6.00 & 22.70 & 4.78 & 4.30 & 7.77 & 12.50 & 7.76 \\
~ & \textbf{PropNet(Ours)} & \textbf{2.96} & \textbf{12.58} & \textbf{2.55} & \textbf{2.44} & \textbf{1.46} & \textbf{5.16} & \textbf{3.75} \\
\hline
\multirow{7}{*}{$\text{AUC}_\text{10\%}$} & ESR \cite{cao2014face} & 0.2774 & 0.0177 & 0.1981 & 0.2953 & 0.2485 & 0.1946 & 0.2204 \\
~ & SDM \cite{xiong2013supervised} & 0.3002 & 0.0226 & 0.2293 & 0.3237 & 0.3125 & 0.2060 & 0.2398 \\
~ & CFSS \cite{zhu2015face} & 0.3659 & 0.0632 & 0.3157 & 0.3854 & 0.3691 & 0.2688 & 0.3037 \\
~ & DVLN \cite{wu2017leveraging} & 0.4551 & 0.1474 & 0.3889 & 0.4743 & 0.4494 & 0.3794 & 0.3973 \\
~ & LAB \cite{wu2018look} & 0.5323 & 0.2345 & 0.4951 & 0.5433 & 0.5394 & 0.4490 & 0.4630 \\
~ & Wing \cite{feng2018wing} & 0.5504 & 0.3100 & 0.4959 & 0.5408 & 0.5582 & 0.4885 & 0.4918 \\
~ & \textbf{PropNet(Ours)} & \textbf{0.6158} & \textbf{0.3823} & \textbf{0.6281} & \textbf{0.6164} & \textbf{0.6389} & \textbf{0.5721} & \textbf{0.5836} \\
\hline
\end{tabular}
\end{table*}

	In order to address data imbalance, we introduce a factor named \textbf{Focal Factor}. For a class $c$ and a sample $n$, it is mathematically defined as:
\begin{equation}\label{factor}
\sigma_n^{(c)} =
\begin{cases}
1 , & \text{if }\sum_{n=1}^{N}{s_n^{(c)}} = 0\\
\frac{N}{\sum_{n=1}^{N}{s_n^{(c)}}} , & \text{otherwise}
\end{cases}
\end{equation}
	where $s_n^{(c)}$ is binary number: when $s_n^{(c)}=0$, the sample $n$ does not belong to class $c$; when $s_n^{(c)}=1$, the sample $n$ belongs to class $c$. In this paper, a sample that belongs to a certain class means the sample has the $c$th attribute, such as large head pose, exaggerated expression, \etc For WFLW dataset, these attributes are labeled in annotation file, while for COFW and 300W we hand-label these attributes by ourselves and use them when training. Also note that the Focal Factor is defined batch-wise, which means it is fluctuating during the training process and again it dynamically adjusts loss weight on every sample in a batch. Furthermore, the weight loss is a sum of all focal factors from different class, as can be seen in the following definition (\ref{lm_loss}). This indicates that we intend to balance the data across all classes because a facial image can be subjected to multiple extreme conditions, \eg a blurry facial image with large head pose.

	As a result, we have the loss of landmark as:
\begin{equation}\label{lm_loss}
L_{lm} = \frac{1}{N}\sum_{n=1}^{N}\sum_{c=1}^C\sigma_n^{(c)}\sum_{k=1}^{K}AWing\left(y_n^{(k)},\hat{y}_n^{(k)}\right)
\end{equation}
	where $N, C, K$ respectively denote batch size, number of classes (subsets) and number of coordinates. In our case, $C=6$ for 6 attributes: head pose, expression, illumination, make-up, occlusion, and blurring; $K=196$ for 98 landmarks that are considered in WFLW dataset. $y_n^{(k)}$ and $\hat{y}_n^{(k)}$ separately stand for the ground truth heatmap of sample $n$, landmark $k$ and the corresponding predicted heatmap.

	Similarly, we define the loss function for boundary heatmap prediction as:
	\begin{equation}\label{bd_loss}
	L_{bd} = \frac{1}{N}\sum_{n=1}^{N}\sum_{c=1}^C\sigma_n^{(c)}\sum_{m=1}^{M}AWing\left(z_n^{(m)},\hat{z}_n^{(m)}\right)
	\end{equation}
	where $M$ denotes the total number of boundaries. $z_n^{(m)}$ and $\hat{z}_n^{(m)}$ are respectively the ground truth boundary heatmap of sample $n$, boundary $m$ and the corresponding predicted boundary heatmap.

	Finally we obtain the holistic loss function as:
	\begin{equation}
	Loss = L_{lm} + \beta \cdot L_{bd}
	\end{equation}
	where $\beta$ is a hyper-parameter for balancing two tasks.

\begin{figure}
\centering
\includegraphics[width=0.48\textwidth]{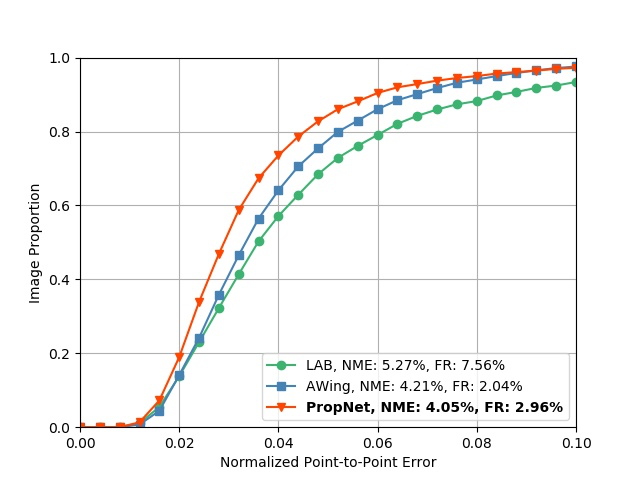} 
\caption{CED for WFLW testset. NME and $\text{FR}_\text{10\%}$ are reported at the legend for comparison. We compare our methods with other state-of-the-arts with source codes available, including LAB \cite{wu2018look} and AWing \cite{wang2019adaptive}.}
\label{CED}
\end{figure}

\subsection{Multi-view Hourglass Module}
	
	Different from traditional hourglass networks using bottleneck block as their building blocks, we adopt the \textbf{hierarchical, parallel and multi-scale residual architecture} proposed by Bulat \etal \cite{bulat2017binarized}. We think the architecture is beneficial to landmark localization due to its multiple receptive fields and the various scale of images those fields can bring. That means we have features describing the larger structure a human face as well as details about each boundary. Hence we name hourglass module as Multi-view Hourglass module and the architecture itself as Multi-view Block, as seen in Fig. \ref{architecture}.

	On the other hands, we implement \textbf{anti-aliased CNN} in place of pooling layer used in traditional hourglass networks. One reason for this is to maintain shift equality in our network while another reason is that we do not want to lose some detail information caused by pooling layer or strided convolution. 

\subsection{Anti-aliased CNN and CoordConv}

	CoordConv \cite{liu2018intriguing} is applied in our work to learn either complete translation invariance or ranging degrees of translation dependence. Anti-aliased CNN \cite{zhang2019making} is also used to replace pooling layer or strided convolution in our work to preserve shift equality. We term it anti-aliased block, seen in Fig. \ref{architecture}.

\section{Experiments}

\subsection{Evaluation Metrics}

	\textbf{Normalized Mean Error (NME)} is a widely used metric to evaluate the performance of a facial landmark localization algorithm. Pixel-wise absolute distance is normalized over a distance that takes face size into consideration. Error of each keypoint is calculated this way and then averaged to get the final result. See Equation (\ref{NME}).
	\begin{equation}\label{NME}
	NME\left(P, \hat{P}\right)=\frac{1}{L}\sum_{l=1}^{L}\frac{\left\|p_l-\hat{p_l}\right\|_2}{d}
	\end{equation}
	where $P, \hat{P}$ are respectively the ground truth coordinates of all points and the predicted ones for a face image, $L$ is the total number of keypoints, and $p_l, \hat{p_l}$ are both 2-dimensional vector presenting the x-y coordinates of the i-th keypoint. Especially, $d$ is the mentioned normalization factor, be it inter-pupil distance or inter-ocular distance. The latter could be distance between the inner corners of eyes (not commonly used) or the outer corner of eyes which we use in our evaluation. For 300W dataset, both factors is applied; for COFW dataset, we use only inter-pupil distance; for WFLW dataset, inter-ocular distance is adopted.

\begin{figure}[t]
\centering
\includegraphics[width=0.48\textwidth]{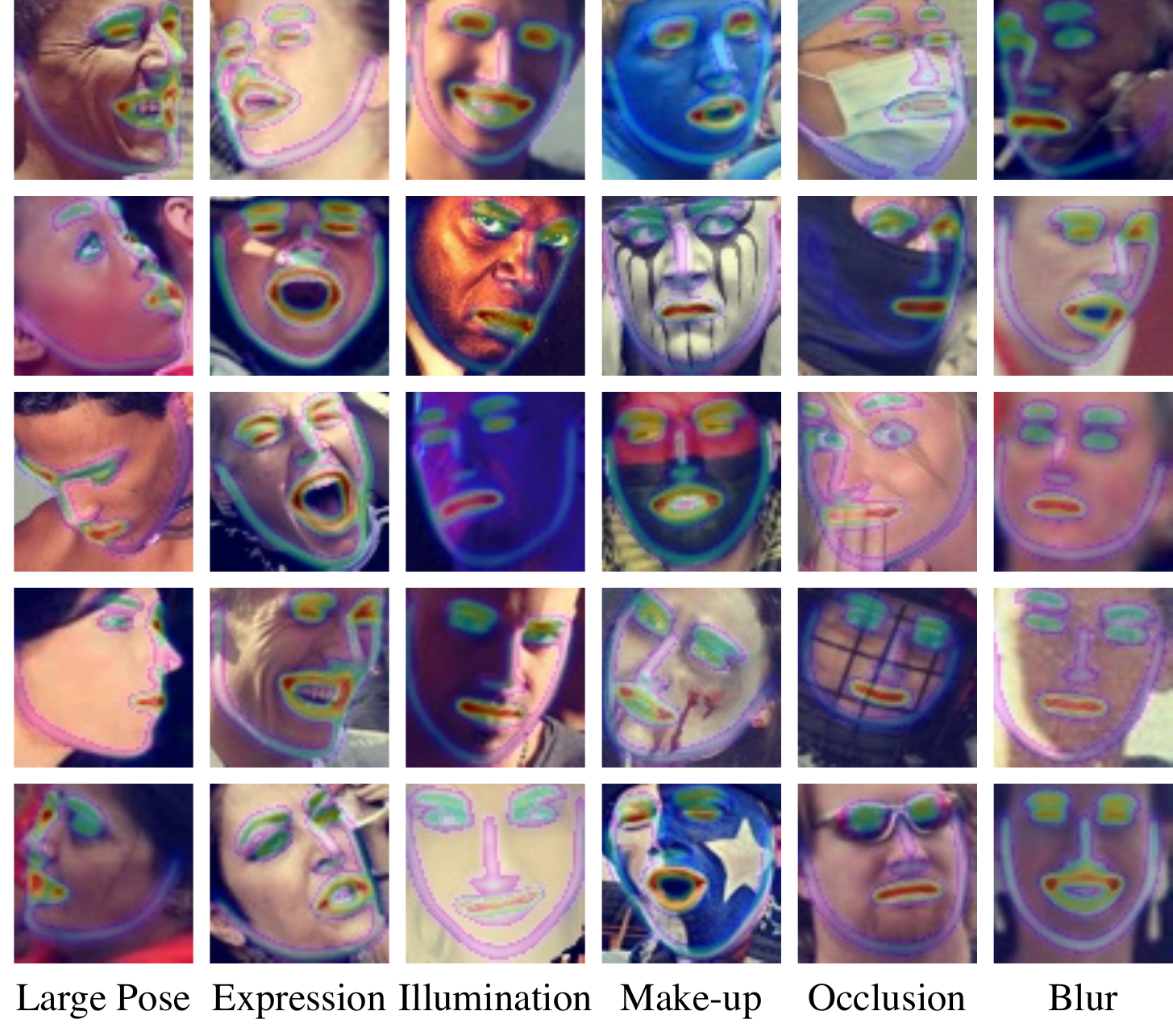} 
\caption{Image samples from WFLW testset imposed with generated boundary heatmap. Each column comes from different subset.}
\label{sample}
\end{figure}

	\textbf{Failure Rate (FR)} provides another insight into a face alignment algorithm design. The NME computed on every image is thresholded at, for example, $8\%$ or $10\%$. If the NME of an image is larger than the threshold, the sample is considered to be a failure. We derive the FR from the rate of failures in a testset.

	\textbf{Area under the Curve (AUC)} is yet another metric popular among designers of facial landmark detection algorithm. Basically, we derive it from the CED curve: by plotting the curve from zero to the threshold for FR, we have a non-negative curve under which the area is calculated to be AUC. An AUC increment implies that more samples in testset are well predicted.

\subsection{Datasets}
	We perform training and testing of our model on 3 datasets: the challenging dataset WFLW \cite{wu2018look} which consists of 10,000 faces (7,500 for training and 2,500 for testing) with 98 fully manual annotated landmarks and is probably hitherto the largest open dataset for face alignment with a large number of keypoints annotation; COFW dataset \cite{burgos2013robust} which contains 1852 face images (1,345 for training and 507 for testing) with 29 annotated landmarks and features heavy occlusions and large shape variations; 300W \cite{sagonas2013300} which is the first facial landmark localization benchmark and for its testset, includes 554 samples for common subset and 135 images for challenging subset.

	On \textbf{WFLW} dataset, we achieve state-of-the-art performance. See Table \ref{WFLW}. Compared with the second leading algorithm, \ie Wing, we improve 3 metrics by about 20\% for NME, around 51\% for $\text{FR}_\text{10\%}$, and roughly 12\% for $\text{AUC}_\text{10\%}$. More importantly, we outperform all the others algorithms on all subsets, which means our model remains robust against different in-the-wild conditions. Special attention should be paid to pose and make-up subsets, where we made a significant improvement. Some sample from the testset can be viewed in Fig. \ref{ResultWFLW}. Besides, we also draw Cumulative Error Distribution (CED) curves (see Fig. \ref{CED}) for algorithms with available released code, including LAB \cite{wu2018look} and AWing \cite{wang2019adaptive}. From the figure, it is obvious that our PropNet curve is higher than the rest two between 0.02 and 0.08, which means we are able to predict facial landmarks of a larger fraction of images in WFLW testset.

	On \textbf{COFW} dataset, our algorithm outperforms the other models. See Table \ref{COFW}. As we all know that COFW is well-known for heavy occlusion and wide range of head pose variation, our leading NME and $\text{FR}_\text{10\%}$ prove that our algorithm stays robust again those extreme situations. This also implies that Propagation Module is able to infuse the network with geometrical structure of a human face because only this structure remains in those worst case scenarios. We can see this in Fig. \ref{ResultW}.

	On \textbf{300W} dataset, our model performs excellently on both two subsets and the fullset when compared with other algorithms using inter-ocular normalization factor, as the upper part of Table \ref{300W} demonstrates. In terms of metrics vis-à-vis inter-pupil normalization, we have similar metrics with the other leading algorithm on the common set and the fullset, but beat them on the challenging set. This suggests that our algorithm can make plausible predictions even in deplorable situations. This is obviously demonstrated in Fig. \ref{ResultW}. A potential reason for relative higher NME with inter-pupil normalization is that 300W annotates some out-of-bounding-box facial parts, \eg chin, with a flat line along the bounding box rather than sticks to the truth. Therefore, this annotation style makes it difficult for our model to learn facial structure.

\subsection{Implementation Details}

	Every input image is cropped and resized to $256\times256$ and the output feature map of every hourglass module is $64\times64$. In our network architecture, we adopt four stacked hourglass modules. During training process, we use Adam to optimize our neural network with the initial learning rate set as $1\times {10}^{-4}$. Besides, data augmentation is implemented at training time: random rotation ($\pm {30}^\circ$), random scaling ($\pm 15\%$), random crop ($\pm 25px$), and random horizontal flip ($50\%$). At test time, we adopt the same strategy of slightly modifying the predicted result as \cite{newell2016stacked}, that is, the coordinate of highest heatmap response is shifted a quarter pixel away to the coordinate of second highest response next to it. Moreover, we empirically set the hyper-parameters in our loss function to be: $\alpha=2.1, \beta=0.5, \omega=14, \epsilon=1.0, \theta=0.5$.

\subsection{Ablation study}

	Our algorithm is comprised of several pivotal designs, \ie Propagation Module (PM), Hourglass Module (HM), and Focal Wing Loss. We will delve into the effectiveness of these components in the following paragraph. For comparison, we use stacked hourglass model with ResNet block as our baseline and it is trained with adaptive wing loss.

\begin{table}[h]
\caption{The potential of Propagation Module's (PM) contribution to our model's performance.}\label{PM}
\centering
\begin{tabular}{|c|c|c|}
\hline
Method & Without PM & With PM  \\
\hline
NME & 4.81 & 4.48 \\
\hline
$\text{FR}_\text{10\%}$ & 3.36 & 3.12 \\
\hline
$\text{AUC}_\text{10\%}$ & 0.5132 & 0.5421 \\
\hline
\end{tabular}
\end{table}

	\textbf{Propagation Module} plays an important role in enhancing our model's performance. It makes the largest improvement to our model. We set our baseline as a stacked hourglass network without this module. See Table \ref{PM} and compare the relation-block-enhanced model with the baseline model. We can observe $-6.86\%$, $-7.14\%$, $5.63\%$ increase respectively in NME (the lower the better), FR (the lower the better), and AUC (the larger the better). From Fig. \ref{sample}, we can see the actual results of generated boundary heatmaps. They are consistent with our expectation, and substantiate our presumption that landmark heatmaps can be propagated to boundary heatmaps via a few consecutive convolution operations. Furthermore, note that our algorithm remain robust in extreme conditions, especially when human face is being occluded, which means the structural information has been captured through our propagation module.

\begin{table}[h]
\caption{The potential of Multi-view Hourglass Module's (MHM) contribution to our model's performance compared to baseline model with Bottleneck Block (BB).}\label{MHM}
\centering
\begin{tabular}{|c|c|c|}
\hline
Method & BB & MHM \\
\hline
NME & 4.81 & 4.67 \\
\hline
$\text{FR}_\text{10\%}$ & 3.36 & 3.16 \\
\hline
$\text{AUC}_\text{10\%}$ & 0.5132 & 0.5300 \\
\hline
\end{tabular}
\end{table}

\begin{table}[h]
\caption{Comparison between anti-aliased CNN (AC) with different size of Gaussian kernel.}\label{AC}
\centering
\begin{tabular}{|c|c|c|c|c|}
\hline
Method & BL & BL+AC-2 & BL+AC-3 & BL+AC-5\\
\hline
NME & 4.81 & 4.79 & 4.67 & 4.75 \\
\hline
$\text{FR}_\text{10\%}$ & 3.36 & 3.80 & 3.16 & 3.76 \\
\hline
$\text{AUC}_\text{10\%}$ & 0.5132 & 0.5178 & 0.5300 & 0.5200\\
\hline
\end{tabular}
\end{table}

	\textbf{Hourglass Module} is one effective module to improve our network's performance on WFLW dataset. Take a look at Table \ref{MHM}. In comparison with the baseline model with bottleneck block, it increases all three metrics by about $-2.91\%$, $-5.95\%$, $3.27\%$. When encountered with the choice of Gaussian kernel size for anti-aliased CNN, we compare different sizes against the baseline model. See Table \ref{AC}. We use $\text{AC-}n$ to indicate the Gaussian kernel of size $n$. For example, $\text{AC-}3$ stands out and we use the size $3$ in the rest experiments.

\begin{table}[h]
\caption{The potential of Focal Wing Loss's contribution to overall performance.}\label{FWL}
\centering
\begin{tabular}{|c|c|c|}
\hline
Method & AWing & Focal Wing\\
\hline
NME & 4.81 & 4.64 \\
\hline
$\text{FR}_\text{10\%}$ & 3.36 & 3.32 \\
\hline
$\text{AUC}_\text{10\%}$ & 0.5132 & 0.5195 \\
\hline
\end{tabular}
\end{table}

	\textbf{Focal Wing Loss} also contributes to the improvement of our model’s performance. As can been seen in Table \ref{FWL}, it gives a rise to three metric increments compared to baseline model trained with AWing by around $-3.53\%$, $-1.19\%$, $1.23\%$ respectively. In addition, we can also see from Table \ref{WFLW} that our model performs better than other state-of-the-arts on every subsets, which means data imbalance has been effectively addressed and once again it helps our network to sustain its robustness against extreme conditions (see fig. \ref{ResultWFLW}).

\begin{table}[h]
\caption{Complexity of PropNet and some other state-of-the-arts.}\label{comp}
\centering
\begin{tabular}{|c|c|c|c|}
\hline
Method & LAB\cite{wu2018look} & AWing\cite{wang2019adaptive} & PropNet \\
\hline
\#params (M) & 12.29 & 24.15 & 36.30\\
\hline
FLOPS (G) & 18.85 & 26.79 & 42.83\\
\hline
\end{tabular}
\end{table}

	See Table \ref{comp}. We make a comparison of computational complexity with some of the open-source state-of-the-arts. As can be seen in the table, we have greater number of parameters and FLOPS than the other two, which may explain why we achieve better performance than them.

\section{Conclusion}

	In our paper, we pinpoint the long-ignored relation between landmark heatmaps and boundary heatmaps. To this end, we propose a Propagation Module to capture the structure information of human face and bridge the gap between landmark heatmap and boundary heatmap. This module is proven by our extensive experiments on widely recognized datasets to be effective and beneficial to the improvement of our algorithm's performance.

	Then we creatively formulate our method to solve data imbalance by introducing the Focal Factor, a factor attempting to dynamically accommodate the loss weight on each sample in a batch. As our ablation studies show, it makes our algorithm more robust under extreme conditions.

	Finally, we also redesign hourglass network by incorporating multi-view blocks and anti-aliased network. The multi-view blocks enables our network to have both macro and micro receptive fields while the anti-aliased architecture make our network shift invariant again. Our ablation studies substantiate its usefulness in the enhancement of our performance.

\section{Acknowledgment}

	This work was sponsored by DiDi GAIA Research Collaboration Initiative.

{\small
\bibliographystyle{ieee_fullname}
\bibliography{reference}
}

\end{document}